\documentclass{article}
\usepackage{ijcai23}

\usepackage{times}
\usepackage{soul}
\usepackage{url}
\usepackage[hidelinks]{hyperref}
\usepackage[utf8]{inputenc}
\usepackage[small]{caption}
\usepackage{graphicx}
\usepackage{amsmath}
\usepackage{amsthm}
\usepackage{booktabs}
\usepackage{algorithm}
\usepackage{algpseudocode}
\usepackage[switch]{lineno}
\usepackage{tablefootnote}
\usepackage{multicol}
\usepackage{tabularx}
\usepackage{multirow}

\title{SuperEdge: Towards a Generalization Model for Self-Supervised Edge Detection}

\author{
    Author Name
    \affiliations
    Affiliation
    \emails
    email@example.com
}
\iftrue
\author{
Kai Leng$^1$
\and
Zhijie Zhang$^1$\and
Jie Liu$^{1}$\and
Wei Sui$^{4}$\and
Zeyd Boukhers$^{3}$\and
Cong Yang$^{2,*}$
\And
Zhijun Li$^{1,2}$\footnote{corresponding: Cong Yang (cong.yang@suda.edu.cn) and Zhijun Li (lizhijunos@hit.edu.cn)}
\affiliations
$^1$School of Computer Science and Technology, Harbin Institute of Technology\\
$^2$School of Future Science and Engineering, Soochow Universit\\
$^3$Fraunhofer Institute for Applied Information Technolog, Sankt Augustin, Germany\\
$^4$Horizon Robotics\\
}
\fi

\begin{document}
\maketitle

\begin{abstract}


Edge detection is a fundamental technique in various computer vision tasks. 
Edges are indeed effectively delineated by pixel discontinuity and can offer reliable structural information even in textureless areas.
State-of-the-art heavily relies on pixel-wise annotations, which are labor-intensive and subject to inconsistencies when acquired manually.
In this work, we propose a novel self-supervised approach for edge detection that employs a multi-level, multi-homography technique to transfer annotations from synthetic to real-world datasets.
To fully leverage the generated edge annotations, we developed SuperEdge, a streamlined yet efficient model capable of concurrently extracting edges at pixel-level and object-level granularity.
Thanks to self-supervised training, our method eliminates the dependency on manual annotated edge labels, thereby enhancing its generalizability across diverse datasets.
Comparative evaluations reveal that SuperEdge advances edge detection, demonstrating improvements of 4.9\% in ODS and 3.3\% in OIS over the existing STEdge method on BIPEDv2.

\end{abstract}
\section{introduction}

Edge detection is a fundamental task in pixel-level computer vision, which involves identifying and extracting discontinuities in images. These discontinuities typically coincide with changes in object features, such as shape, color, material, and other attributes. Playing an important role in complex vision tasks like semantic segmentation\cite{GRSL_2023_EDGN}, depth estimation\cite{IROS_2022_AERO}, and object detection\cite{TNNLS_2021_LAB}, edge detection has been extensively researched. Nonetheless, it continues to pose challenges and remains open to innovative contributions.
\begin{figure}[h]  
    \centering  
    \includegraphics[width=0.48\textwidth]{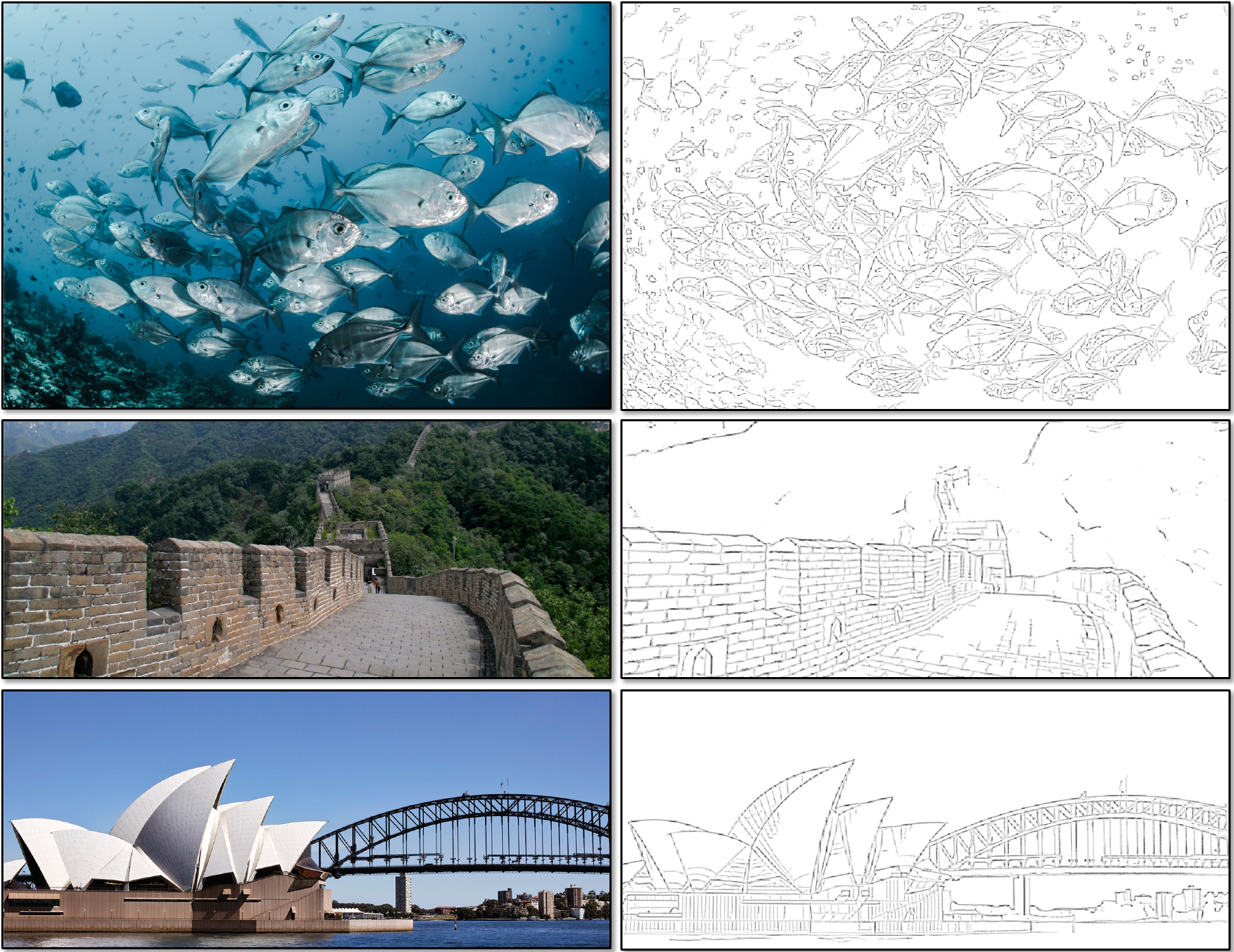}  
    \caption{The edge-maps predictions from the proposed model in images acquired from internet.} 
    \label{fig:internet}
    \vspace{-1.5em}
\end{figure}

Traditional method uses local features based on gradient and filter to predict the edge\cite{ESEX_2010_SURVEY}, such as Canny\cite{TPAMI_1986_CANNY}.
Recently, learning-based edge detection methods have shown promising progress. Efficient algorithms such as HED\cite{ICCV_2015_HED}, BCDN\cite{CVPR_2019_BCDN}, DexiNed\cite{WCAV_2020_DexiNed}, PidiNet\cite{ICCV_2021_PiDiNet} and TEED\cite{ICCV_2023_TEED} have been developed. These algorithms typically undergo training on manually annotated datasets and incorporate multi-layer supervision to enhance the models' generalization capabilities.

A notable drawback of existing edge detection methodologies is their substantial reliance on manually annotated data, which is both costly and labor-intensive. Furthermore, edge annotation suffers from a critical issue of inconsistency, where different annotators often provide divergent interpretations of edge boundaries within the same image. This is in contrast to tasks such as object detection, where annotations tend to be more consistent. For instance, the classical edge detection dataset BSDS\cite{ICCV_2001_BSDS} was annotated by five individuals, and the Multicue\cite{VR_2016_Multicue} dataset involved six annotators, leading to frequent discrepancies. Consequently, the ground truths are typically obtained by taking the average of these annotations.


Based on the above observations, we propose an edge self-labeling methodology for transferring from virtual synthetic scenes to real scenes. This self-supervised generation method is based on Homography Adaption proposed by SuperPoint\cite{CVPR_2018_SuperPoint}, which we adapt and apply to edge detection, marking its inaugural use in this domain. Similar to keypoints, this method of annotation generation exhibits high sensitivity to pixel-level edges within images. However, it falls short in capturing object-level edges. To address this limitation, we draw inspiration from STEdge\cite{TNNLS_2023_STEdge} and employ L0-smoothing\cite{TPAMI_2011_L0SM} for image clustering based on color attributes, in conjunction with the Canny algorithm, to extract object-level edges effectively.

Additionally, we propose a simple but effective edge detector named SuperEdge. This model adopts an encoder-decoder architecture, distinguishing itself from prior approaches that simultaneously optimize predictions across all convolutional layers before computing the mean. Our model strategically bifurcates its decoder output heads into two distinct streams: one dedicated to extracting pixel-level edges and the other to object-level edges. This design aligns seamlessly with the self-labeled data produced by our method and eliminates the complexity of concurrently optimizing multiple convolutional layers.
 When trained on self-supervised data from COCO2017, our strategy and model outperform the state-of-the-art method STEdge across multiple datasets. Furthermore, our approach demonstrates effective edge extraction capabilities in complex scenes (refer to Fig.\ref{fig:internet}).

Succinctly, the main contributions are as follows:
\begin{itemize}
    
    \item We introduce a self-supervised strategy for edge detection that effectively generates pixel-level and object-level edge annotations in real-world scenes, obviating the need for labor-intensive manual labeling.
    \vspace{-0.3em}
    \item Based on the generated data, we designed a novel model named SuperEdge. Extensive evaluations of the self-supervised strategy and model demonstrate its strong cross-dataset generality.
    \vspace{-0.3em}
\end{itemize}
\section{Related Work}

\textbf{Edge detection methods} can be simply divided into three categories: traditional edge detectors, learning-based detectors and deep learning-based edge detectors. 

Traditional methods include techniques such as Sobel\cite{SU_1970_sobel} and Canny\cite{TPAMI_1986_CANNY}, which extract edges by directly performing gradient analysis on the image. Learning based methods encompasses approaches that amalgamate diverse low-level features and employ prior knowledge to train detectors capable of generating edges. 
Deep learning-based edge detection predominantly employs convolutional neural network (CNN) architectures, with prominent methods including HED\cite{ICCV_2015_HED}, DexiNed\cite{WCAV_2020_DexiNed}, PiDiNet\cite{ICCV_2021_PiDiNet}. These methods typically undergo training on manually annotated datasets designed for edge detection and frequently incorporate techniques such as multi-layer regularization and image augmentation to bolster the method's generalization capabilities. For a more comprehensive review, we refer readers to \cite{NC_2022_survey}.

\textbf{Datasets}
is another  pivotal role in deep learning-based edge detection methodologies. Current edge detection datasets include BSDS\cite{ICCV_2001_BSDS}, Multicue\cite{VR_2016_Multicue} and NYUD\cite{ECCV_2012_NYUD}, etc. Often, the introduction of a new edge detection approach is accompanied by the release of a new dataset tailored for the method. For instance, RindNet\cite{ICCV_2021_RindNet} not only classifies edges in images but also further annotates the BSDS dataset to create the enhanced BSDS-RIND dataset. Similarly, alongside DexiNed, the BIPED dataset was introduced, followed by an extended version, BIPEDv2\cite{PR_2023_BIPEv2}, which provides further annotations for images in BIPED. The primary motivation for the continuous introduction of new datasets is to improve model generalization capabilities. However, the creation of these datasets is an arduous and labor-intensive process, typically necessitating the efforts of multiple annotators. Furthermore, the quantity of manually labeled data remains substantially limited, with most datasets comprising only several hundred images for training and a handful for testing. This quantity is insufficient for deep learning-based methods, prompting the utilization of various data augmentation techniques, such as rotation and scaling, to increase the number of the training data. Addressing the challenges associated with manual data annotation, STEdge\cite{TNNLS_2023_STEdge} introduced a novel self-supervised edge detection strategy. It employs multi-layer regularization in conjunction with the L0-smoothing algorithm and the Canny operator, facilitating self-supervised model training through iterative loops. This innovation enables the model to initially reduce its reliance on manually labeled datasets. Nevertheless, the heavy dependence on the Canny operator for iteration sometimes leads to noticeably noisy predictions in many scenarios.

\section{Method}
\begin{figure*}
    \centering  
    \includegraphics[width=\textwidth]{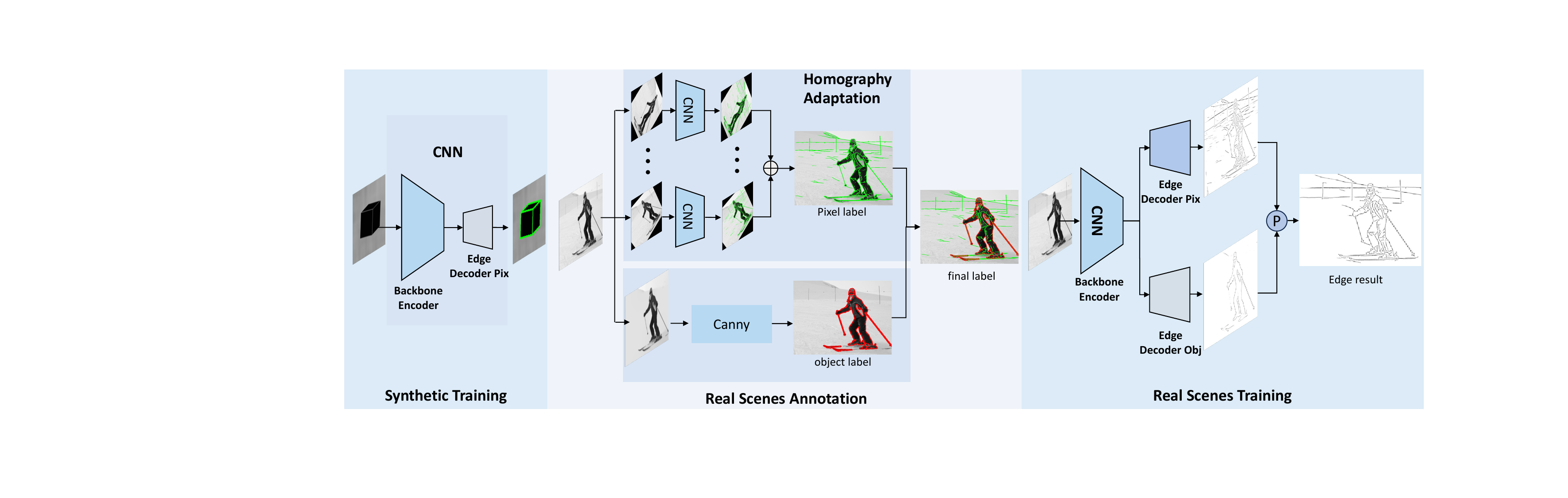}  
    \vspace{-1.5em}
    \caption{\textbf{Training pipeline overview.} \textbf{Left:} Our detector is first trained on an edge synthetic dataset of known ground truth. 
    \textbf{middle:} Object-level and pixel-level pseudo ground truth of edge are generated on real images through methods such as L0 smoothing, canny, and homography adaptation.
    \textbf{Right:} Finally, the SuperEdge is trained on real images using the pseudo ground truth.
    } 
    \label{fig:overview}
    \vspace{-1.0em}
\end{figure*}

\label{sec:method}
\subsection{Overview}
We propose an edge detector based on the encoder-decoder architecture that can effectively extract pixel-level edges and object-level edges within images. Our entire self-supervised training strategy unfolds in three steps:  initially, our model is trained on a synthetic edge dataset with known ground truth. Subsequently, leveraging the model pre-trained on synthetic data, we perform edge annotation on real-world scenes, utilizing methods such as homography adaptation, L0-smoothing, and the Canny algorithm. Finally, we train SuperEdge on real images using pseudo labels. Fig.\ref{fig:overview} provides an overview of our self-supervised pipeline, which we expound upon later in this section.
\subsection{Self-supervise strategy}
Inspired by the success of SuperPoint\cite{CVPR_2018_SuperPoint} in keypoint detection and SOLD$^2$\cite{CVPR_2021_Pautrat} in line detection. 
We extend their homography adaptation to the field of edge detection to realize the self-annotation of edges in real-world scenes.

In the \textbf{Synthetic Training} stage, we train on a virtual synthetic dataset containing known edge annotations. As shown in Fig.\ref{fig:overview} left, the synthetic dataset is generated by introducing different geometric shapes (such as cubes, polygons, and stars, etc.) under the background of random Gaussian noise.


In the \textbf{Real Scene Annotation} stage, consider the model pre-trained on synthetic data as a function $f(\cdot)$. Let $I$ denote the input image and $\mathcal{H}$ an arbitrary homography transformation. We introduce a multitude of random homography transformations to the process. By aggregating the outcomes from these diverse transformations, we synthesize a more robust edge predictor denoted as $F(\cdot)$. This enhanced predictor not only excels within simulated environments but also demonstrates commendable accuracy and stability when applied to real-world scenes. This underpins the effective transition from synthetic to real-world datasets. Based on these settings, we can get the following formula:
\begin{equation}
F\left(I ; f\right)=\frac{1}{N_h} \sum_{i=1}^{N_h} \mathcal{H}_i^{-1} f\left(\mathcal{H}_i(I)\right) \quad .
\end{equation}
where $N_h$ represents the number of homography adaptations applied, and $\mathcal{H}_i^{-1}$ is the inverse transformation corresponding to the $i$-th homography, ensuring that the predictions are aligned with the original image domain.

Utilizing the annotator function $F(\cdot)$, our framework is capable of extracting pixel-level edges from images, with intermediate results depicted in Fig.\ref{fig:overview} middle. 
This approach is adequate when extracting pixel-level edges is the sole objective. Yet, to enhance the granularity of edge processing, it is imperative to consider the object-level edges. 
 we have engineered a process specifically to annotate object-level edges within images, denoted as the function $G(\cdot)$. This process is structured into the following sequential steps:
 \begin{enumerate}

\item Application of Gaussian blur followed by L0-smoothing to reduce noise and enhance image smoothness, aiding in the demarcation of object boundaries.
\item Deployment of the Canny edge detection algorithm, succeeded by an image dilation process to highlight object-level edges.
\end{enumerate}

The culmination of these procedures results in the precise and enriched annotation of object-level edges. Then, the final generation of self-supervised annotation can be expressed as:
\begin{equation}
x = F(I,f) + G(I)\quad .
\end{equation}
where $x$ represents the combined edge annotations.

Finally, \textbf{in the Real Scenes Training} stage, we trained the final model, SuperEdge, using the annotations generated from real-world scenes as described in the second stage.
\begin{figure*}
    \centering  
    \vspace{-1.5em}
    \includegraphics[width=\textwidth]{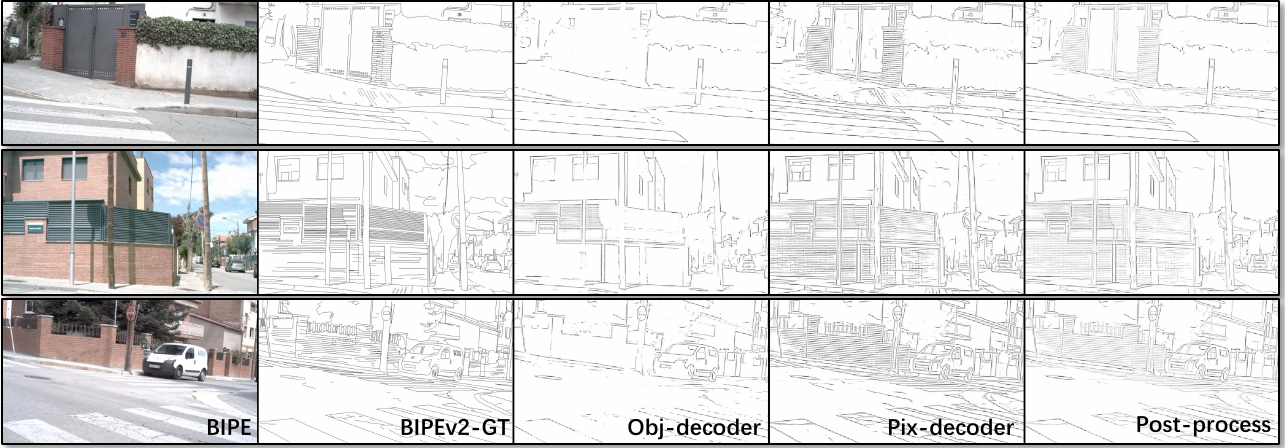}  
    \caption{Results of each detection head of SuperEdge on BIPED.
    } 
    \vspace{-1.0em}
    \label{fig:ablation_view}
\end{figure*}
\subsection{Model and Loss design}
\begin{table}[!ht]
    \caption{SuperEdge overview, where \textit{B} represents batchsize, \textit{H} and \textit{W} are the height and width of the input image, respectively.} 
    \vspace{-1em}
    \label{tab:superedge_overview}
    \begin{tabularx}{0.49\textwidth}{cccc}
        \toprule
         \multicolumn{4}{c}{Overview} \\
         \multicolumn{2}{c}{\makebox[0.23\textwidth]{Operation}}  & \multicolumn{2}{c}{\makebox[0.22\textwidth]{Output Shape}} 
         \\
    
        \midrule
        \multicolumn{2}{l}{\textbf{Input Image}} & \multicolumn{2}{l}{[\textit{B}, \textit{H}, \textit{W}]} \\
        \multicolumn{2}{l}{\textbf{Encoder}} & \multicolumn{2}{l}{[\textit{B}, \textit{128}, \textit{H/8}, \textit{W/8}]} \\
        \multicolumn{2}{r}{\textit{vgg block1}} & \multicolumn{2}{r}{[\textit{B}, \textit{64}, \textit{H/2}, \textit{W/2}]} \\
        \multicolumn{2}{r}{\textit{vgg block2}} & \multicolumn{2}{r}{[\textit{B}, \textit{64}, \textit{H/4}, \textit{W/4}]} \\
        \multicolumn{2}{r}{\textit{vgg block3}} & \multicolumn{2}{r}{[\textit{B}, \textit{128}, \textit{H/8}, \textit{W/8}]} \\
        \multicolumn{2}{r}{\textit{vgg block4}} & \multicolumn{2}{r}{[\textit{B}, \textit{128}, \textit{H/8}, \textit{W/8}]} \\
        \multicolumn{2}{l}{\textbf{Pixel-edge decoder}} & [\textit{B}, \textit{65}, \textit{H/8}, \textit{W/8}] \\
        \multicolumn{2}{r}{\textit{conv2d, bn, relu}} & \multicolumn{2}{r}{[\textit{B}, \textit{256}, \textit{H/8}, \textit{W/8}]} \\
        \multicolumn{2}{r}{\textit{conv2d, bn, softmax}} & \multicolumn{2}{r}{[\textit{B}, \textit{65}, \textit{H/8}, \textit{W/8}]} \\
        \multicolumn{2}{r}{\textit{reshape}} & \multicolumn{2}{r}{[\textit{B}, \textit{H}, \textit{W}]} \\
        \multicolumn{2}{l}{\textbf{Object-edge decoder}} & [\textit{B}, \textit{65}, \textit{H/8}, \textit{W/8}] \\
        \multicolumn{2}{r}{\textit{conv2d, bn, relu}} & \multicolumn{2}{r}{[\textit{B}, \textit{256}, \textit{H/8}, \textit{W/8}]} \\
        \multicolumn{2}{r}{\textit{conv2d, bn}} & \multicolumn{2}{r}{[\textit{B}, \textit{195}, \textit{H/8}, \textit{W/8}]} \\
        \multicolumn{2}{r}{\textit{self-attention, softmax}} & \multicolumn{2}{r}{[\textit{B}, \textit{65}, \textit{H/8}, \textit{W/8}] }\\
        \multicolumn{2}{r}{\textit{reshape}} & \multicolumn{2}{r}{[\textit{B}, \textit{H}, \textit{W}]} \\
        \bottomrule
    \end{tabularx}
    \vspace{-1em}
\end{table}
As shown in Fig.\ref{fig:overview} right and Tab.\ref{tab:superedge_overview}, our model uses grayscale images as input and performs feature extraction through a shared backbone encoder. In order to correspond to the self-annotation generated by the Real Scene Annotation stage, we split the output into two branches. One branch is used to predict pixel-level edges, while the other branch focuses on detecting object-level edges. 

For the pixel-level branch, we adopt a similar approach to the SuperPoint keypoint extraction method. 
The output of the pixel decoder yields a coarse feature map $\mathbf{P}$ of size $\frac{W}{8} \times \frac{H}{8} \times 65$. Each 65-dimensional vector corresponds to an $8 \times 8$ patch along with an additional "no edge" dustbin.
Subsequently, the pixel-level edge loss is computed as the cross-entropy loss between feature map $\mathbf{P}$ and ground truth $\mathbf{y}$, as shown in Eq.\ref{loss_pix}.
The aforementioned loss function was initially designed for keypoint extraction using the SuperPoint model. Its original purpose was to decompose the image into 64 sub-images, with the model predicting the channel to which each pixel position belongs. However, given the possibility of multiple key points within each patch, it has been observed that satisfactory predictions can still be achieved. Considering this phenomenon, we have applied loss function $\mathcal{L}_{pix}$ to the task of pixel-level edge detection:
\begin{equation}
\label{loss_pix}
\mathcal{L}_{pix}= \frac{64}{H \times W} \sum_{\substack{h=1, w=1}}^{\frac{H}{8},\frac{W}{8}}{l}_{p}(x_{hw};y_{hw})
\quad .
\end{equation}
where
\begin{equation}
{l}_{p}(x_{hw};y)= -\log \left(\frac{\exp \left(\mathbf{P}_{hwy}\right)}{\sum_{k=1}^{65} \exp \left(\mathbf{P}_{hwk}\right)}\right)
\quad .
\end{equation}
where $H$ and $W$ are the length and width of the input image. In order to obtain the final $H \times W $ grid for the edge map, we apply a softmax operation along the channel dimension, followed by removing the 65th dimension.

For the object-level branch, considering that the task necessitates the acquisition of more semantic information, we introduce a single-layer self-attention mechanism to allow the model to better obtain global information to understand the semantics and structure of the entire image. 
Given the substantial imbalance in the distribution of edge/non-edge pixels within natural images, we employ the robust weighted cross-entropy loss, denoted as $\mathcal{L}_{obj}$, to guide the training process. The loss is computed by:
\begin{equation}
\mathcal{L}_{\mathrm{obj}}= 
\begin{cases}
\alpha \cdot{l}_{p}(x_{hw};y_{hw}), & \text { if } y_{hw} =65 \\
\beta \cdot {l}_{p}(x_{hw};y_{hw}), & \text { if } y_{hw} < 65 
\end{cases}
\quad .
\end{equation}
in which
\begin{equation}
\begin{aligned}
& \alpha=\lambda \cdot \frac{\left|Y^{+}\right|}{\left|Y^{+}\right|+\left|Y^{-}\right|} \\
& \beta=\frac{\left|Y^{-}\right|}{|Y^{+}|+|Y^{-}|}
\end{aligned}
\quad .
\end{equation}
where $Y^{+}$ and $Y^{-}$ represent the counts of edge pixels and non-edge pixels in the pseudo ground truth, respectively. $\lambda$ is a hyper-parameter to balance positive and negative samples. Finally, our loss function is defined as follows:
\begin{equation}
    \mathcal{L}_{sum} = \mathcal{L}_{pix} + \mathcal{L}_{obj}
    \quad .
\end{equation}
\subsection{Post-process module}
As shown in Fig.\ref{fig:ablation_view}, the pixel-level edge detector can effectively extract various types of edges in the image. However, due to the influence of noise and other unstable factors, pixel discontinuities may not truly reflect the actual edges of the image. Object-level edges can effectively reduce noise in prediction results, but they lack a lot of detailed information. Therefore, we proposed a simple post-process algorithm. This algorithm uses object-level edge prediction results as a skeleton, which is obtained by continuously expanding the pixel-level edge prediction results based on Breadth-first search (BFS). See Appendix A for the specific algorithm. The visualization results are shown on the right of Fig. \ref{fig:ablation_view}. The final result is obtained as follow:
\begin{equation}
    O_{fusion} = norm(\frac{\mathcal{P}(O_{pix},O_{obj}) + O_{obj}}{2} )
    \quad .
\end{equation}
where $O_{pix}$ and $O_{obj}$ are the edge prediction results of two detection heads, respectively. $\mathcal{P}$ is the post-processing operation. $norm$ is the image normalization operation.

\subsection{Implementation details}
The source code is available in \url{https://github.com/lktidaohuoxing/SuperEdge}. To optimize the network, we employ the Adam optimizer, with the learning rate set to 0.001. Each training session consists of 100 epochs, utilizing a batch size of 16. In practice, the training process on the $120 \times 160$ Synthetic dataset is completed within approximately 1 hour. Subsequently, we proceed to label and train on the $240 \times 320$ COCO-train2017\cite{ECCV_2014_coco} dataset, which requires approximately two days. Following that, secondary labeling and training on the $480 \times 640$ COCO-val2017 dataset are performed, taking a total of approximately 10 hours. Additionally, we set both the pixel-level edge threshold and the object-level edge threshold to 0.005 for optimal performance.

\section{Experiments}

\subsection{Dataset}
\label{sec:dataset}

The datasets utilized in our experimental analysis encompass COCO\cite{ECCV_2014_coco}, BSDS500\cite{ICCV_2001_BSDS}, BSDS-RIND\cite{pu_2021_rindnet}, BIPED\cite{WCAV_2020_DexiNed} and BIPEDv2\cite{PR_2023_BIPEv2}. 

BSDS dataset is specifically created for image segmentation and edge detection tasks, comprising 200 training set samples, 100 validation set samples, and 200 test set samples. Each image in BSDS is annotated by a minimum of 5 annotators, making it a valuable resource for image segmentation and boundary detection research.
BSDS-RIND dataset is a further extension of the BSDS dataset. It categorizes edges within images into four distinct types: reflectance edges, illumination edges, normal edges, and depth edges. This additional annotation enriches the content and diversity of the dataset.
BIPED dataset consists of 250 annotated images of outdoor scenes, divided into a training set comprising 200 images and a testing set containing 50 images. BIPEDv2 represents the final iteration of the BIPED dataset, incorporating further annotations and refinements.
 NYUD dataset comprises 1449 RGBD images, encompassing 464 indoor scenarios, specifically tailored for segmentation purposes. Following \cite{gupta_2013_perceptual}, this dataset is subdivided into training, validation, and testing subsets, with 654 images reserved for testing, while the remaining images are allocated for training and validation purposes.

\subsection{Metric}
As with previous work on edge detection, we use the following three indicators to evaluate the effect of model:
\begin{enumerate}
    \item Optimal Dataset Scale (ODS) computed by using a global threshold for the entire dataset.
    \vspace{-0.3em}
    \item Optimal Image Scale (OIS) computed by using a different threshold on every image.
    \vspace{-0.3em}
    \item Average Precision (AP) is the integral of the precision/recall (P/R) curve. 
    \vspace{-0.3em}
\end{enumerate}


It should be noted that in some cases, the P/R curve is shorter and does not cover the entire range, in this case
AP values are lower and the assessment reliability of AP may be lower compared to ODS and OIS. Additionally, in our experiments, we followed the convention of setting the maximum allowable distance between the predicted edge and the corresponding pixel of the ground truth value to 0.0075 for both ODS and OIS evaluations.
\subsection{Ablation study}
As mentioned in Sec. \ref{sec:method}, we introduce a network model based on an encoder-decoder architecture, incorporating two decoders responsible for pixel-level edges and object-level edge detection. In order to verify the effectiveness of each module in this design, we specifically conducted a series of ablation experiments. These experiments were designed to independently evaluate the contribution of each decoder head in the edge detection task, as well as the impact of the post-processing algorithm on the results. 

\begin{table}[htbp]
\vspace{-0.5em}
\caption{Ablation experiments on BIPED dataset} 
\vspace{-0.5em}
\label{tab:SuperEdge_ablation_study}
\renewcommand{\arraystretch}{1} 
\centering
    \begin{tabular*}{0.48\textwidth}{@{\extracolsep{\fill}}lccc}
        \toprule
        Method & ODS & OIS & AP \\
        \midrule
        object-decoder & 0.707  & 0.707 & 0.537\\

        pixel-decoder & 0.784  & 0.789 & 0.680\\
        pixel\&object  & 0.787 & 0.799 & \textbf{0.758} \\
        pixel\&object + post-process  & \textbf{0.811} & \textbf{0.818} & 0.755\\
        \bottomrule
    \end{tabular*}
\end{table}

\begin{figure}
    \centering  
    \includegraphics[width=0.45\textwidth]{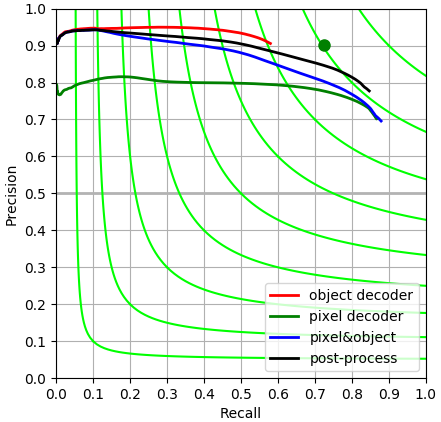}  
    \vspace{-0.5em}
    \caption{\textbf{P/R  curves} on BIPED dataset.
    } 
    \vspace{-1.2em}
    \label{fig:ablation_PR}
\end{figure}

The ablation experiments were performed on the BIPED dataset, evaluating the results obtained from the two detection heads individually, their combination through simple addition, and the utilization of a post-processing algorithm. The quantitative results and corresponding P/R curves are presented in Tab. \ref{tab:SuperEdge_ablation_study} and Fig. \ref{fig:ablation_PR}, respectively. Additionally, visualization results are depicted in Fig. \ref{fig:ablation_view}.

As indicated in Tab. \ref{tab:SuperEdge_ablation_study}, the AP values of the individual output results from the two detection decoders are relatively low. However, when combined through a simple addition, the overall AP value shows a significant improvement. This observation suggests the effective complementarity between the two detection decoders for different types of edges.
Furthermore, the introduction of the post-processing algorithm leads to noticeable improvements in ODS and OIS, even when the AP values remain comparable. This demonstrates that the post-processing algorithm is highly effective in eliminating edge noise at the pixel level. It is clear from the quantitative results table and the visual results comparison chart that each decoder head and post-processing algorithm plays an important role in the edge detection task.
This further confirms the correctness and effectiveness of SuperEdge's design idea.

\subsection{Comparison study}

\begin{table}[ht]
    \centering
        \vspace{-1.0em}
    \caption{Comparative experiments on BIPED dataset. }
    \vspace{-0.5em}
    \begin{tabular}{ccccc}
    \toprule
        Method & Train & ODS & OIS & AP \\
        \midrule
        HED & BSDS & 0.711 & 0.725 & 0.714 \\
        BDCN & BSDS  & 0.714 & 0.725 & 0.687 \\
        DexiNed & BSDS  & 0.699 & 0.716 & 0.664 \\
        PiDiNet & BSDS  & 0.776 & 0.785 & 0.740 \\
        \midrule
        DexiNed-ST & COCO-val2017*  & 0.760 & 0.783 & \textbf{0.798}\\
        PiDiNet-ST & COCO-val2017* & 0.786 & 0.804 & 0.259 \\
        \midrule
        SuperEdge & COCO-val2017*  & \textbf{0.811} & \textbf{0.818}  & 0.755 \\
        \bottomrule
    \end{tabular}
    \label{tab:BIPE_compare}
        \vspace{-0.5em}
\end{table}

\begin{table*}[h]
    \caption{Comparative experiments on different datasets, all models were trained on the COCO-val2017.}
    \label{tab:multi_compare}
    \centering
    \renewcommand{\arraystretch}{1.5} 
    
    \begin{tabular}{cccc|ccc|ccc|ccc}\toprule
        &\multicolumn{3}{c}{BIPEDv2} & \multicolumn{3}{c}{BSDS-RIND} & \multicolumn{3}{c}{NYUD} &\multicolumn{3}{c}{BSDS500}    \\
        \cline{2-4}
        \cline{5-7}
        \cline{8-10}
        \cline{11-13}
        & ODS & OIS & AP & ODS & OIS & AP & ODS & OIS & AP & ODS & OIS & AP \\
        \midrule
        DexiNed-ST & 0.780 & 0.804 & 0.668 & 0.662 & 0.685 & 0.544 & 0.536 & 0.553 & 0.385 & 0.648 & 0.659 & 0.527 \\
        PiDiNet-ST & 0.786 & 0.802 & 0.340 & 0.670 & 0.695 & 0.223 & 0.544 & 0.556 & 0.114 & 0.656 & 0.672 & 0.186 \\
        SuperEdge & \textbf{0.825} & \textbf{0.829} & \textbf{0.752} & \textbf{0.722} & \textbf{0.733} & \textbf{0.640} & \textbf{0.581} & \textbf{0.591} & \textbf{0.438} & 
        \textbf{0.672} & \textbf{0.686} & \textbf{0.571} \\
        \bottomrule
    \end{tabular}
\end{table*}
\begin{figure*}
    \centering  
    \includegraphics[width=\textwidth]{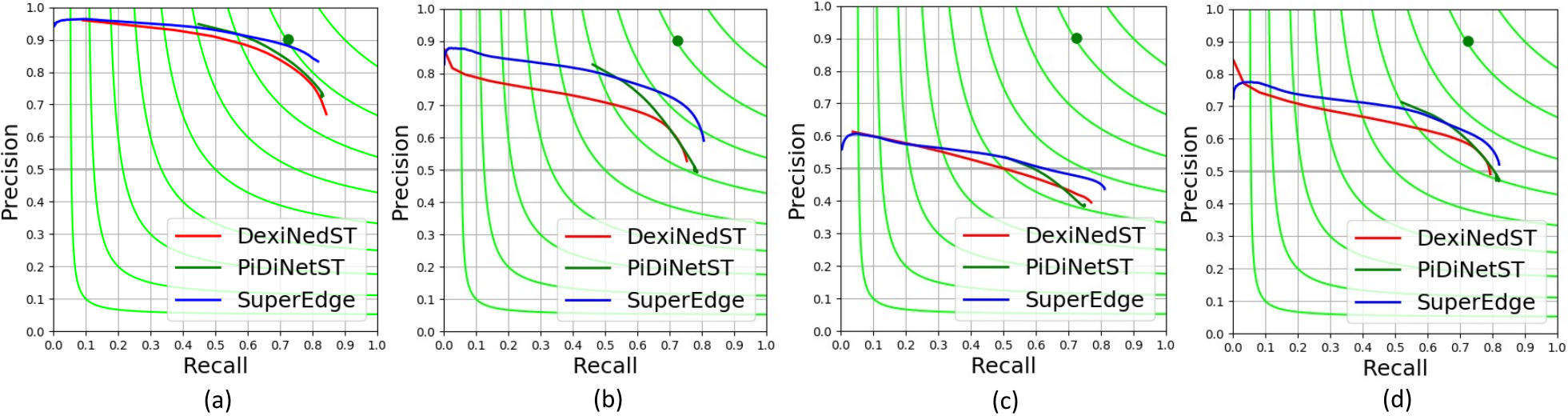}  
    \vspace{-1.8em}
    \caption{P/R curves on different datasets. (a) BIPEDv2. (b) BSDS-RIND. (c) NYUD. (d) BSDS500.
    } 
    \vspace{-1.0em}
    \label{fig:compare_multi_pr}

\end{figure*}
We compare SuperEdge with existing self-supervised method STEdge\cite{TNNLS_2023_STEdge} across multiple models, including PidiNet\cite{ICCV_2021_PiDiNet} and DexiNed\cite{WCAV_2020_DexiNed}, which are two lightweight edge detection methods. We also compared it against classical methods such as HED\cite{ICCV_2015_HED} and BDCN\cite{CVPR_2019_BCDN}.

First, we trained four methods, namely HED, BDCN, and PiDiNet on the BSDS dataset. Subsequently, we retrained DexiNed and PiDiNet using the self-supervised strategy implemented in STEdge on the COCO-val2017 dataset, resulting in DexiNed-ST and PiDiNet-ST. Finally, this study applied the self-supervised data generation method proposed in Sec. \ref{sec:method} to train SuperEdge on the COCO-val2017 dataset. The quantitative results are presented in Tab. \ref{tab:BIPE_compare}.
Upon observing and comparing the results, we discovered that the models trained using the STEdge self-supervised strategy exhibited superior generalization capabilities on unseen data. Specifically, they achieved higher ODS and OIS scores on the untouched BIPED dataset. Furthermore, compared to other self-supervised comparative methods, our model and training approach demonstrated significant advantages. Notably, there was a 3.1\% improvement in ODS and a 1.7\% improvement in OIS performance. These results validate the effectiveness of the self-supervised method and SuperEdge model proposed in this study for complex edge detection tasks.

\begin{figure*}[htbp]
    \centering  
    \includegraphics[width=\linewidth]{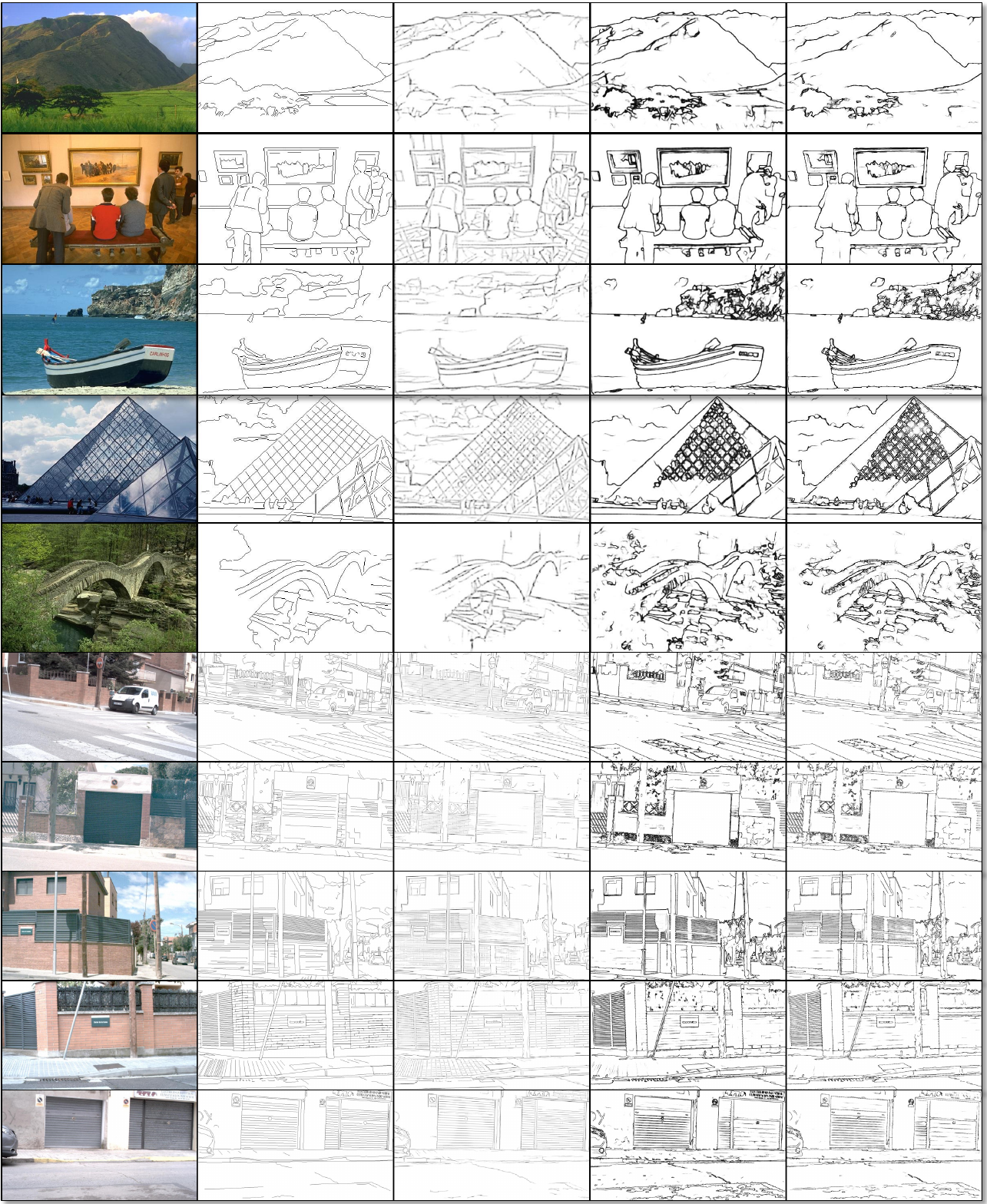} 
    \includegraphics[width=1.0\linewidth]{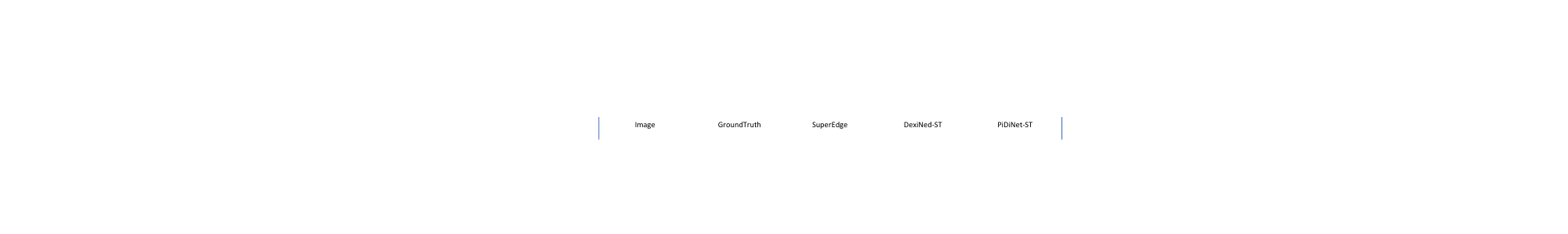}
    \caption{Visualization results on BSDS-RIND (first five rows)  and BIPEDv2 (last five rows). } 
    \label{fig:visual_BIPE_result}
\end{figure*}

Next, this study conducts a more extensive comparison of the performance of SuperEdge, DexiNed-ST, and PiDiNet-ST. We select multiple datasets, including BIPEDv2, BSDS-RIND, BSDS500, and NYUD, which are described in detail in Sec. \ref{sec:dataset}. 
 As shown in the quantification results Tab. \ref{tab:multi_compare} and the corresponding PR curve Fig.\ref{fig:compare_multi_pr}, SuperEdge consistently outperforms other methods across all datasets and metrics, which once again verifies the superiority of the self-supervised method and model design employed in this study for edge detection.  In addition, the visual results of the BIPEDv2 and BSDS-RIND are shown in Fig. \ref{fig:visual_BIPE_result}. Analyzing the visualization results shows that the performance improvement of SuperEdge mainly lies in its robustness to noise (such as lawn, asphalt road, ocean, etc.) and its ability to effectively extract various edges in the image (such as illumination changes, depth changes, object contours, etc.).

\vspace{-0.8em}
\section{Conclusion}

\noindent \textbf{Contribution}.
In this paper, we present a novel self-supervised edge detection method along with its corresponding model design. We introduce Homography Adaptation into the field of edge detection for the first time and combine it with technologies such as Canny and L0-smooth to enable edge annotation in any scene. Furthermore, we propose a dual decoder model called SuperEdge for training in real-world scenarios. Compared with current state-of-the-art methods, SuperEdge achieves huge performance improvements on different datasets.

\noindent \textbf{Limitation}.
Currently, while our method demonstrates strong performance in terms of generalization, the training effect of self-supervision still cannot exceed that of supervised training on specific datasets. Furthermore, the existing search-based post-processing algorithms are computed using CPU resources, thereby underutilizing the potential of GPU resources and resulting in decreased inference speed.

\noindent \textbf{Futrue work}. 
In future research, our goal is to integrate key point detection with edge detection, enabling simultaneous output from a single model. Furthermore, we will explore a unified approach that combines key points and edges to reveal both the geometric representation and edge description of any edge in an image. Our ultimate objective is to extract, describe, and match various elements within an image.

\newpage
\bibliographystyle{named}
\bibliography{ijcai23}

\end{document}